\documentclass{article}

\usepackage{PRIMEarxiv}

\usepackage[utf8]{inputenc} 
\usepackage[T1]{fontenc}    
\usepackage{url}            
\usepackage{booktabs}       
\usepackage{amsfonts}       
\usepackage{nicefrac}       
\usepackage{microtype}      
\usepackage{lipsum}
\usepackage{fancyhdr}       
\usepackage{graphicx}       
\graphicspath{{media/}}     

\usepackage[utf8]{inputenc}
\usepackage{fourier} 
\usepackage{array}
\usepackage{makecell}
\usepackage{xcolor}
\usepackage{varwidth} 
\usepackage[hidelinks]{hyperref}

\title{Better Reasoning Behind Classification Predictions with BERT for Fake News Detection}

\author{
  Daesoo Lee \\
  Norwegian University of Science and Technology \\
  Alfred Getz' vei 1, 7034 Trondheim, Norway \\
  \texttt{daesoo.lee@ntnu.no} \\
}

\begin{document}
\maketitle

\begin{abstract}
Fake news detection has become a major task to solve as there has been an increasing number of fake news on the internet in recent years. Although many classification models have been proposed based on statistical learning methods showing good results, reasoning behind the classification performances may not be enough. In the self-supervised learning studies, it has been highlighted that a quality of representation (embedding) space matters and directly affects a downstream task performance. In this study, a quality of the representation space is analyzed visually and analytically in terms of linear separability for different classes on a real and fake news dataset. To further add interpretability to a classification model, a modification of Class Activation Mapping (CAM) is proposed. The modified CAM provides a CAM score for each word token, where the CAM score on a word token denotes a level of focus on that word token to make the prediction. Finally, it is shown that the naive BERT model topped with a learnable linear layer is enough to achieve robust performance while being compatible with CAM.
\end{abstract}


\section{Introduction}
In recent years, deceptive content such as fake news has become one of the major issues as such fake news can easily be spread out to millions and billions of people through various online platforms. Although there are many reliable news organizations, there are also a large number of smaller news sources that may deliver news that are not trustworthy. In addition, now that anybody can publish any sort of statement on social media platforms, the issue with fake news has become even larger. For instance, the U.S. presidential election in 2016 triggered a major public concern in the United States about the spread of fake news on social media \cite{grinberg2019fake}. 

In recent years, many fake news detection methods have been proposed based on statistical learning methods \cite{ahmed2018detecting, nasir2021fake, seetharaman2022analysis, kaliyar2021fakebert}, in which some are based on traditional methods and others are based on deep learning methods. Those fake news detection methods consist of three main components: 1) tokenization, 2) vectorization, and 3) classification model. The tokenization involves pre-processing such as splitting a sentence into a set of words, removal of the stop words, and stemming. The vectorization is typically achieved by a word count method such as term frequency-inverted document frequency (TF-IDF) or a pre-trained word embedding matrix by unsupervised word embedding models such as Word2Vec \cite{mikolov2013efficient}, GloVe \cite{pennington2014glove}, or BERT \cite{devlin2018bert}. Although those studies \cite{ahmed2018detecting, nasir2021fake, seetharaman2022analysis, kaliyar2021fakebert} have shown gradual improvement in the fake news detection performance over years, the studies are only about experimenting with different models and different pre-trained word embedding matrices, and report which combination works the best. Also, their evaluations are sorely based on classification metrics such as false-positive rate, false-negative rate, and accuracy.

The recent studies in self-supervised learning (SSL) have shown importance of representation space quality \cite{grill2020bootstrap, chen2021exploring, zbontar2021barlow, bardes2021vicreg, lee2021vibcreg}. The SSL methods are designed to train a model in a self-supervised learning manner so that it can model the representation space effectively. One of the commonly-used representations in the SSL methods is the vector after the global-average-pooling (GAP) layer in ResNet \cite{he2016deep}. The representation space refers to the space of the vector. In the \textit{well-fitted} representation space, even a linear layer is enough to do a downstream task such as a classification task robustly. It is commonly known that performance of an NLP model is better with Word2Vec or GloVe than with TF-IDF, and performance with BERT is better than with GloVe or Word2Vec. Therefore, an initial assumption in this study was that a quality of the representation (embedding or vector) space should be ordered as follows (in descending order): BERT, Word2Vec or GloVe, and TF-IDF. To verify the assumption, analysis of a quality of the representation (embedding) space is conducted in this study.

Another limitation in the previous studies is a lack of a method to discover underlying patterns of real and fake news. Pattern discovery is an important aspect of a fake news detection method to answer the following question - "What makes fake news fake and real news real?". In this study, a simple method is proposed to discover underlying patterns of real and fake news based on an interpretation method in computer vision, Class Activation Mapping (CAM) proposed by \cite{zhou2016learning}.

In summary, the main contributions of this study are as follows: 1) Analysis of quality of the representation (embedding) space by TF-IDF, Word2Vec, GloVe, and BERT with respect to classification performance, 2) Proposal of modified CAM to discover underlying patterns of real and fake news, 3) Achievement of both high classification accuracy and interpretability with a simple BERT-based architecture (\textit{i.e.,} naive BERT topped with a linear layer).

\section{Background and Related Work}

\begin{figure}[ht!]
\centering
\includegraphics[width=0.95\textwidth]{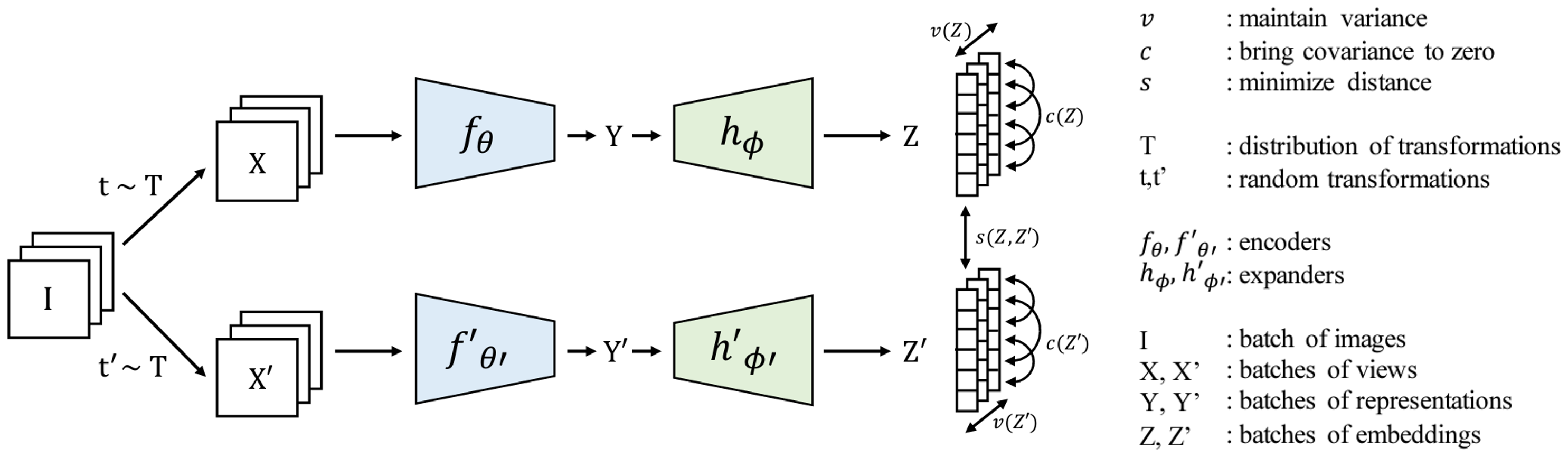}
\caption{Overall framework of one of the recent SSL methods, VICReg. $\mathrm{Y}$ is denoted as \textit{representation} and its space is termed representation space. As for the encoder $f_{\theta}$, ResNet is used. Note that this SSL framework is proposed to effectively fit the representation space in a self-supervised learning manner so that a downstream task such as classification can easily be done with a light prediction head with fine-tuning. \cite{bardes2021vicreg}}
\label{fig:vicreg}
\end{figure}

\begin{figure}[ht!]
\centering
\includegraphics[width=1.0\textwidth]{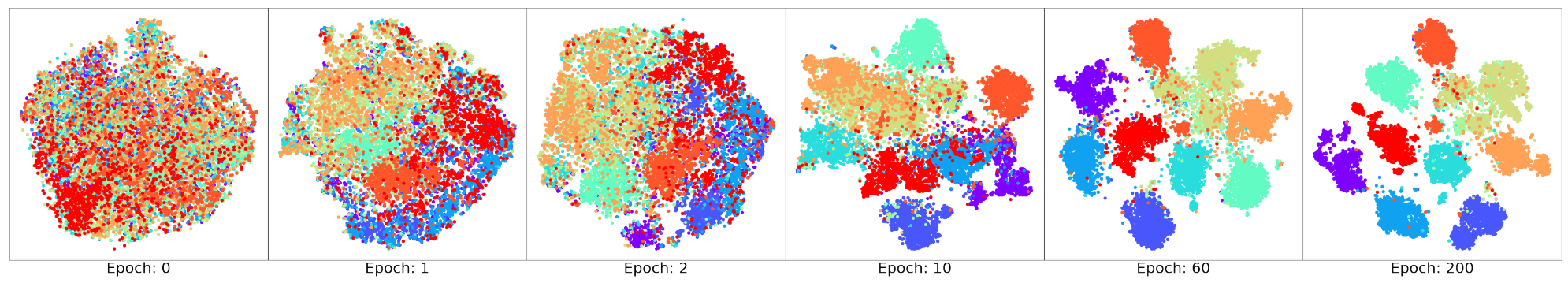}
\caption{Visualization of the learned representations by MSF with 10 different classes on the ImageNet dataset at different epochs. The different colors represent different classes. It can be observed that the representation space is effectively fitted such that samples with the same class are clustered while samples with different classes are pushed away. \cite{koohpayegani2021mean}}
\label{fig:msf_tsne}
\end{figure}

\paragraph{Representation Space} The representation space commonly refers to a vector space where the vector is usually the one right after an encoder such as ResNet. One illustration for the representation space is presented is Fig. \ref{fig:vicreg}, which is originally from a SSL paper that proposed Variance-Invariance-Covariance Regularization (VICReg) \cite{bardes2021vicreg}. After the representation space is fitted by VICReg, the authors discarded all the modules except the encoder, froze the encoder, and added a linear layer onto the frozen encoder. They trained the linear layer only on the ImageNet dataset for a classification task to see how much the learned representations are linearly separable (\textit{i.e.,} higher classification would mean the representations are linearly well separable by the linear layer, meaning the quality of the learned representations is good). They achieved the test accuracy of 73.2\% and 91.1\% for the top-1 and top-5 accuracies, respectively. Another recent SSL study that proposed Mean Shift (MSF) presented nice visualization of the representation space by t-SNE over epochs trained on the ImageNet dataset as Fig. \ref{fig:msf_tsne} \cite{koohpayegani2021mean}.

\begin{figure}[ht!]
\centering
\includegraphics[width=0.62\textwidth]{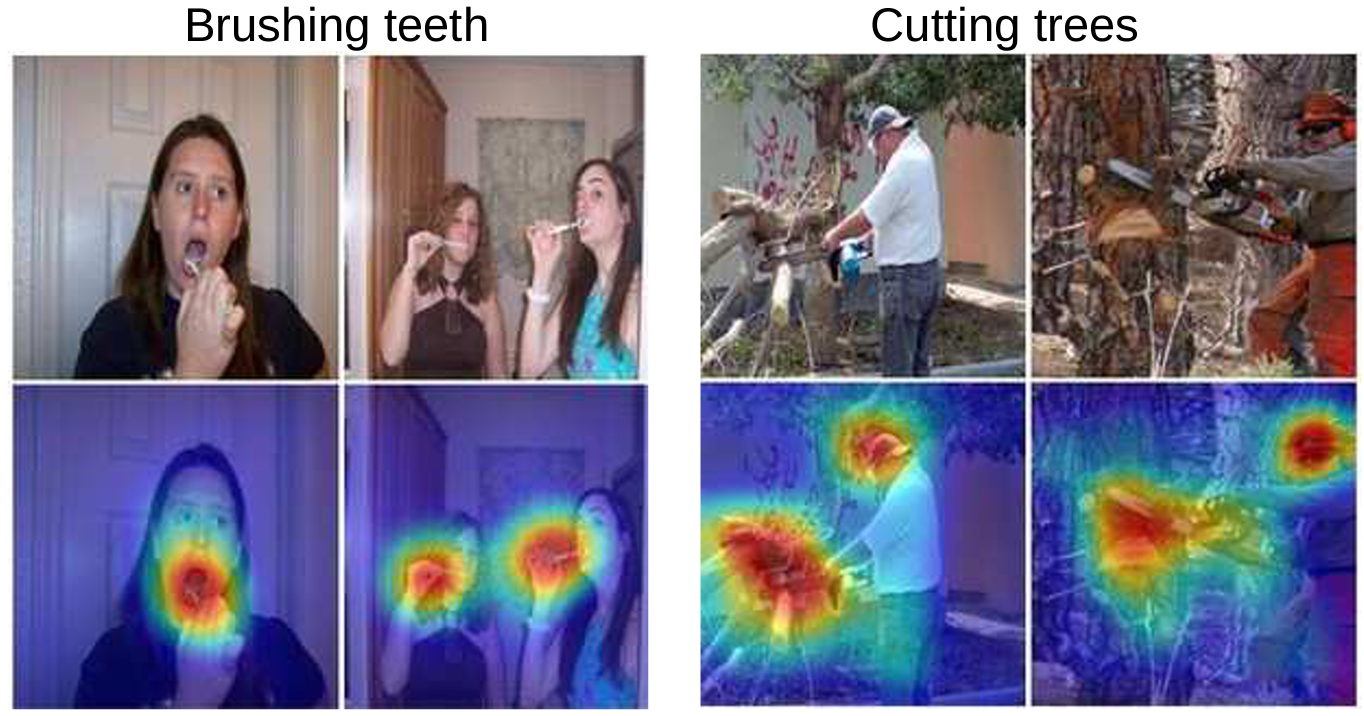}
\caption{Examples of CAM. The strength of the color denotes a level of focus by the model. The red color indicates that the model is putting a strong focus on that part to classify an image as "Brushing teeth" or "Cutting trees" while the blue color indicates a near-zero focus. \cite{zhou2016learning}}
\label{fig:cam_example}
\end{figure}

\paragraph{Interpretability} Deep learning models have been criticized for being a black box for many years. To resolve that issue, many interpretation methods have been proposed for different purposes. One of the most influential interpretation methods from computer vision is CAM \cite{zhou2016learning} and Grad-CAM \cite{selvaraju2017grad}. Some examples of CAM are shown in Fig. \ref{fig:cam_example}. Though CAM was proposed in computer vision, the initial assumption in this study was that it should be possible to apply this to an NLP model with a bit of modification to a model architecture.

\paragraph{Previous Fake News Detection Methods} 

A traditional framework to solve NLP-classification problems consists of the followings: 1) preprocessing a dataset (tokenizing, stop word removal, stemming, etc.), 2) word vectorization, 3) training a classifier. \cite{ahmed2018detecting} followed this traditional approach, where TF-IDF with different n-gram sizes are used for word vectorization and support vector machine, K-nearest neighbor, linear regression, and decision tree are used for the classifier. In a recent year, \cite{nasir2021fake} used a CNN-LSTM architecture, where CNN and LSTM stand for convolutional neural network and long sort term memory, respectively. In their work, a pre-trained GloVe is used for word vectorization and the word vectors are first processed by a CNN layer and then by an LSTM layer, where the CNN captures local dependency while the LSTM captures long-term dependency. In the same year, \cite{kaliyar2021fakebert} proposed to utilize a pre-trained word embedding matrix by BERT with a CNN-based model architecture. They did comparative experiments between GloVe and BERT and showed that BERT's word embedding is more effective than GloVe's.

\paragraph{Word Embedding Models}

\begin{figure}[ht!]
\centering
\includegraphics[width=0.75\textwidth]{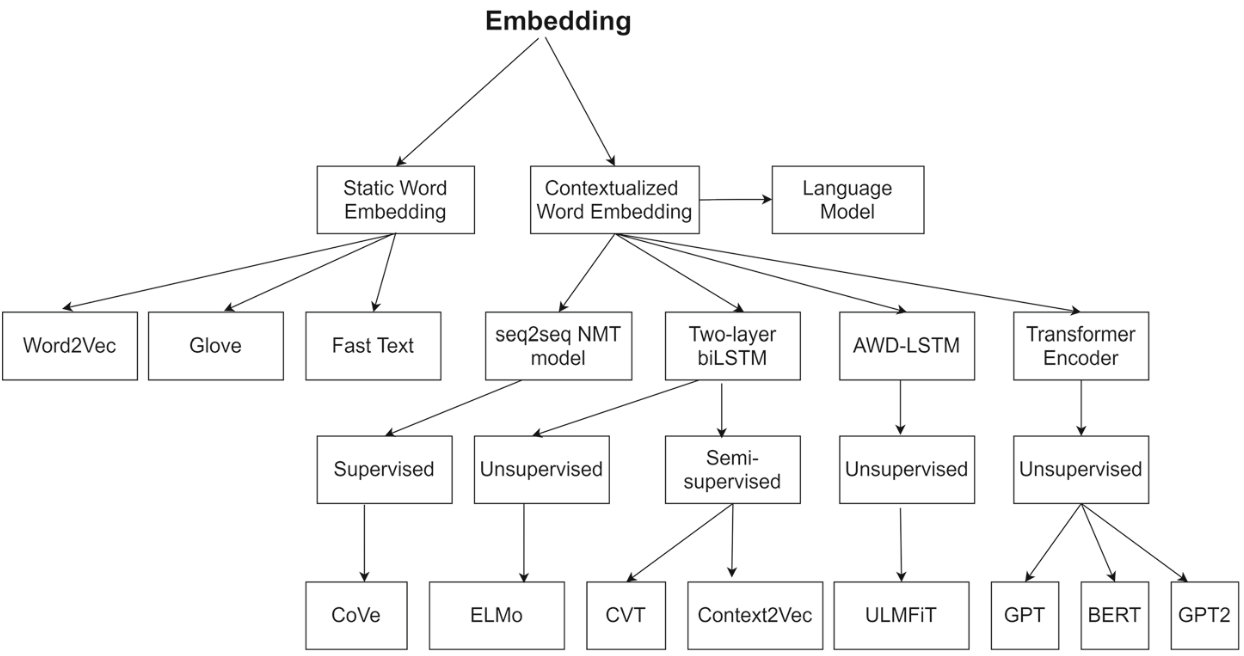}
\caption{Overview of some of the well-known word embedding models \cite{kaliyar2021fakebert}.}
\label{fig:word_emb_models}
\end{figure}

As shown in Fig. \ref{fig:word_emb_models}, types of word embedding models can be divided into two: 1) static word embedding, 2) contextualized word embedding. The main difference is that a word embedding vector by the contextualized model reflects information of other words in the same sequence. Therefore, each word vector by the contextualized model is more context-aware. In this study, some popular embedding models from the static word embedding and contextualized word embedding are considered: Word2Vec, GloVe, and BERT.

\section{Methodology}

\subsection{Models}
\label{sec:architecture}

\begin{figure}[ht!]
\centering
\includegraphics[width=1.0\textwidth]{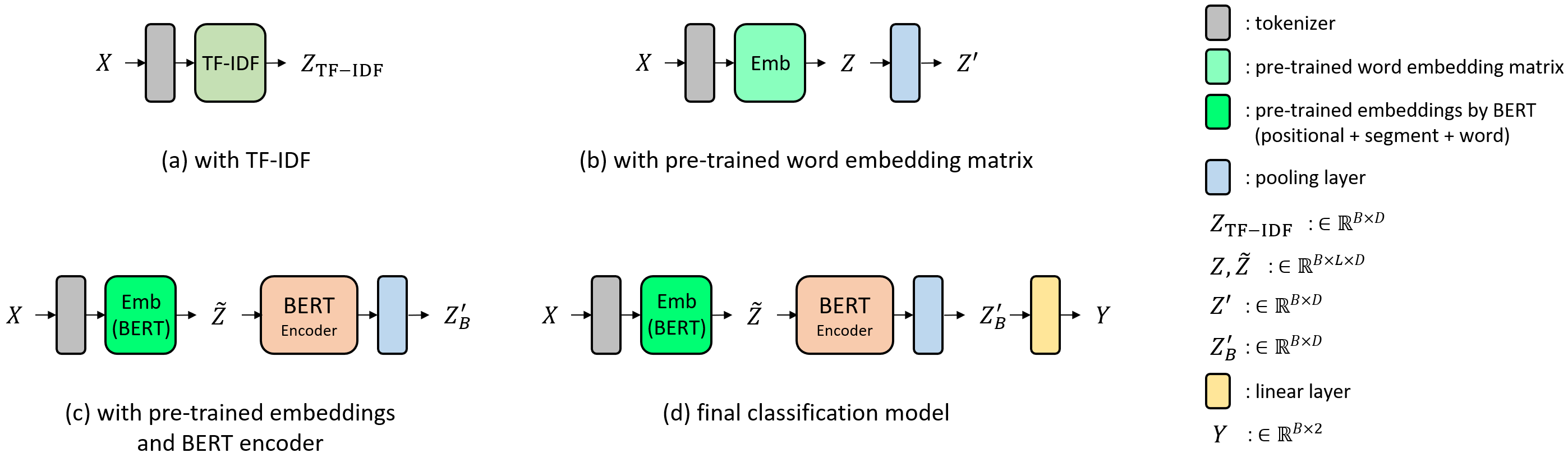}
\caption{(a), (b), and (c) are for analysis of quality of the representation (\textit{i.e.,} static or contextualized word vector) space. (d) is the final classification model used in this study. $X$, $B$, $D$, $L$ denote a batch of sentences, batch size, dimension size, and token length, respectively. $Z_{\mathrm{TF-IDF}}$ denotes a batch of feature vectors from TF-IDF, $Z$ denotes a batch of word vectors from a pre-trained word embedding matrix by Word2Vec, GloVe, or BERT, and $\tilde{Z}$ denotes a batch of word vectors with the pre-trained embeddings by BERT. As for the pooling layer, the GAP layer along the $L$ dimension is used. In the final classification model, $Y$ has vector size of 2 for the two classes: real and fake. Note that cross-entropy loss is used to train the classification model.}
\label{fig:frameworks}
\end{figure}

Fig. \ref{fig:frameworks} presents models for analysis of a quality of the representation (\textit{i.e.,} static or contextualized word vector) space and the final classification model for fake news detection. In Fig. \ref{fig:frameworks}(a), $Z_\mathrm{TF-IDF}$ does not carry any semantics in its feature vector nor any temporal dependency. In Fig. \ref{fig:frameworks}(b), $Z$ carries relational semantics that captures relations between different words, which would make $Z$ a better quality representation than $Z_\mathrm{TF-IDF}$. However, because $Z$ is reduced by the GAP layer, $Z^{\prime}$ does not encode any temporal dependency. In Fig. \ref{fig:frameworks}(c), $\tilde{Z}$ not only involves the word embedding but also the positional embedding and the segment embedding, and it is attended by the BERT's transformer encoder, where the attention mechanism effectively encodes global temporal dependency, then processed by the GAP layer. Therefore, $Z^\prime_{B}$ carries not only the relational semantics but also encodes the global temporal dependency. Hence, it can be intuitively thought that a quality of the representations would be listed as follows: $Z^\prime_{B}$, $Z^\prime$, and $Z_\mathrm{TF-IDF}$. Finally, the final classification model used for the fake news detection is presented in Fig. \ref{fig:frameworks}(d).

\subsection{CAM-based Interpretation Method}
In the original paper of CAM, CAM is applied to a CNN model followed by GAP and linear layers. To make our model compatible with the CAM's framework, the GAP layer and the linear layer are followed after the BERT encoder as shown in Fig. \ref{fig:frameworks}(d). In the original CAM, the equation for the CAM score is defined as Eq. (\ref{eq:cam_original}). $M_c(x,y)$ is the class activation map for class $c$ and it indicates the importance of the activation at spatial grid $(x,y)$ leading to the classification of an image to class $c$. $f_k(x,y)$ represents the activation of unit $k$ (\textit{i.e.,} $k$-th channel of the activation map) in the last convolutional layer at spatial location $(x,y)$. $w_k^c$ is the weight in the linear layer corresponding to class $c$ for unit $k$. Essentially, $w_k^c$ indicates the importance of $\sum_{x,y}f_k(x,y)$ for class $c$. In this study, the spatial dimension does not lie in $(x,y)$ but in the token length $L$. Here, $l$ is used to indicate an element in the $L$ dimension. Then, Eq. (\ref{eq:cam_original})'s notation can be modified to suit our model as Eq. (\ref{eq:cam_mod}). As for the implementation, everything remains the same except that it addresses the $l$ spatial dimension instead of the $(x,y)$ spatial dimension and the $D$ dimension is addressed instead of the channel dimension.

\begin{equation} \label{eq:cam_original}
M_c(x,y) = \sum_{k} w_k^c f_k(x,y)
\end{equation}
\begin{equation} \label{eq:cam_mod}
M_c(l) = \sum_{k} w_k^c f_k(l)
\end{equation}

\section{Experiments and Results}
\label{sec:Experiments}

\subsection{Experimental Setup}
\label{sec:experimentalSetup}

\paragraph{Dataset} 
A dataset named "fake and real news dataset" from Kaggle is used. The dataset can be found here: \url{https://bit.ly/36hZJog}. The dataset was collected from real-world sources. The real news articles were collected from \url{Reuters.com} (News website), and the fake news articles were collected from unreliable websites that Poitifact (a fact-checking organization in the USA) has stamped out. The dataset is class-balanced, in which 23,481 samples for the real news and 21,417 samples for the fake news. Most of the articles are political news articles that were published in 2016. In order not to make the dataset trivial for either classification or representation space analysis, the \textit{text} columns is used as input sentences and "..(Reuters) -" is removed from the real news articles. The dataset is split into a training set and a test set by 80\% and 20\%, respectively.

\paragraph{Tokenizers and Embedding Matrices}
For the tokenizers, there are three types as shown in Fig. \ref{fig:frameworks}. For the framework in Fig. \ref{fig:frameworks}(a), a simple tokenizer included in \texttt{TfidfVectorizer} from the scikit-learn \cite{scikit-learn} is used. or the framework in Fig. \ref{fig:frameworks}(b), three types of pre-trained embedding matrices exist: Word2Vec, GloVe, and BERT. In case of the Word2Vec and GloVe embedding matrices, a simple tokenizer by \texttt{nltk.tokenize} is used. In case of the BERT embedding matrix, \texttt{BertTokenizer} from the huggingface \cite{wolf-etal-2020-transformers} is used. For the embedding matrices by Word2Vec, GloVe, and BERT, \texttt{word2vec-google-news-300}, \texttt{glove-wiki-gigaword-300} from the Gensim \cite{Rehurek_Software_Framework_for_2010}, and \texttt{BertModel} from the huggingface are used, respectively.

\paragraph{Training Settings} 
For an optimizer, AdamW \cite{loshchilov2017decoupled} (\textit{lr}=0.001, batch size=128, weight decay=\{0.1, 0.001, 0.00001\}, epochs=\{1, 2\}). The values within the curly brackets indicate that the best parameter among them is used to report the result. For a learning rate shceduler, a cosine learning rate decay is used \cite{loshchilov2016sgdr}. 

To build and train the deep learning models, PyTorch \cite{paszke2019pytorch} is used with a single GPU (RTX 3060) in this study.

\paragraph{Analysis of Representation Space: PCA} 
The following representation spaces are analyzed in this study: 1) $Z_\mathrm{TF-IDF}$, 2,3,4) $Z^\prime$ with the word embedding matrices of Word2Vec, GloVe, BERT, 5) $Z^\prime_B$. Note that $Z^\prime$ belongs to the static word embedding and $Z^\prime_B$ belongs to the contextualized word embedding. 
In case of TF-IDF, it is fitted on the training set and the representations are obtained on the test set and projected onto 2-dimensional space by principal component analysis (PCA) with the first two principal components (PCs). 
In case of using the pre-trained embedding(s), the representations are obtained on the test set without prior fitting process and projected on the 2-dimensional PC space.
If a representation space quality is good, the representations would be visibly separable according to different classes on the PC space. Typically, t-SNE is used for a visualization analysis tool, PCA is used here to linearly project the representations on 2-dimensional space instead of a non-linear projection method such as t-SNE because our goal is to see if the representations are linearly separable.

\paragraph{Analysis of Representation Space: Linear Evaluation}
The linear evaluation is a conventional evaluation method for SSL methods, which is used to evaluate how much learned representations are linearly separable so that a quality of learned representations can be measured. The linear evaluation is conducted on Fig. \ref{fig:frameworks}(a,b,c). The \textit{linear evaluation protocol} used in this study is as follows: First, all the learnable parameters are frozen in a current model if there is any. Then, a learnable linear layer is added onto the model, fitted on the training set, and evaluated on the test set by test accuracy. The higher test accuracy would indicate that learned representation space is more linearly separable for the different classes.

\subsection{Results and Discussion}
\label{sec:results_discussion}

\begin{figure}[ht!]
\centering
\includegraphics[width=1.0\textwidth]{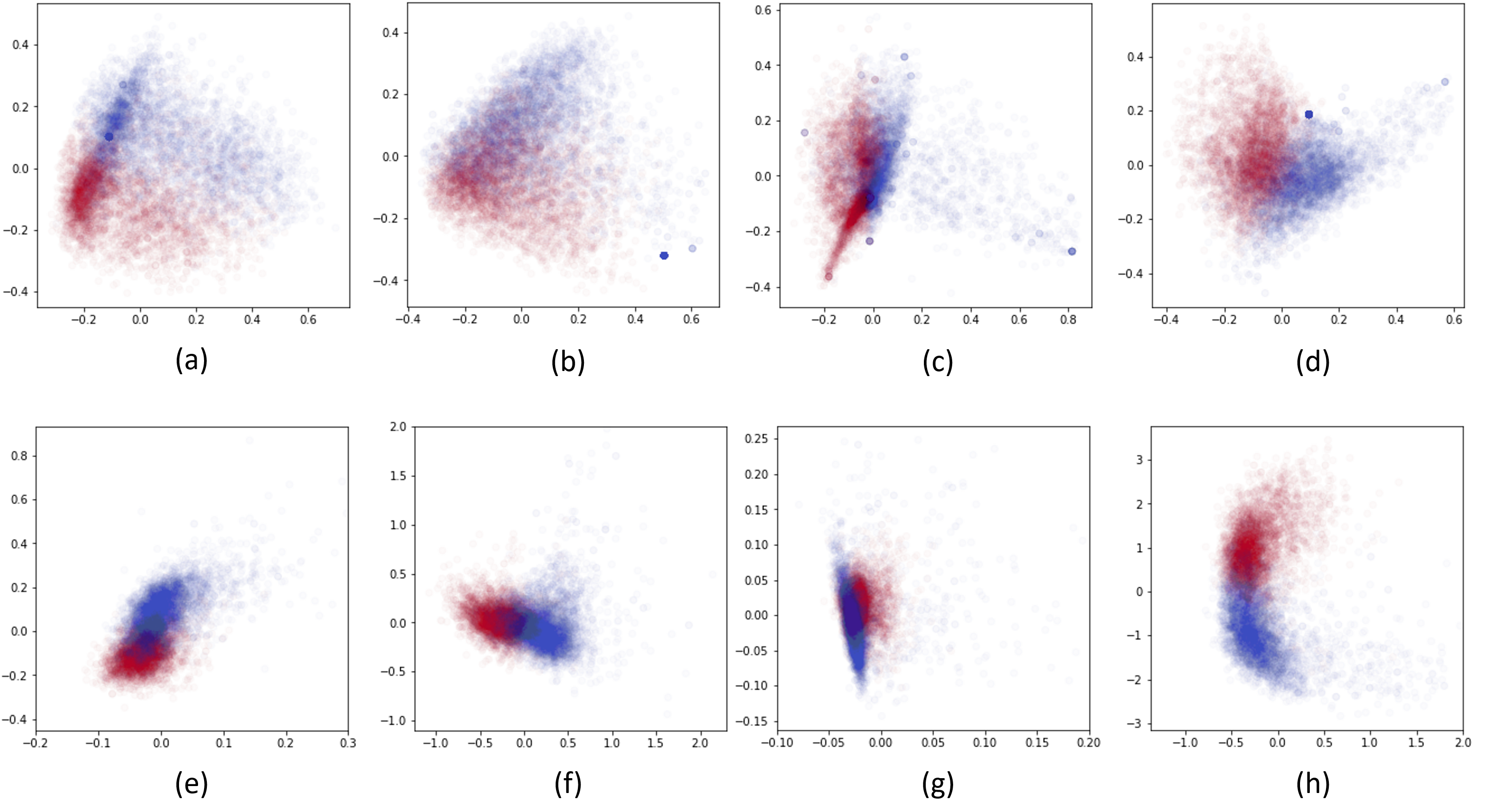}
\caption{Visual analysis of the representation spaces by PCA. Notation of each sub-figure is explained in Table \ref{tab:cases_LE}. The blue and red colors denote the fake and real labels, respectively. Among all the cases, (h) seems to be most linearly separable in a sense that samples are well-clustered with the small overlapping area.}
\label{fig:pca_results}
\end{figure}

\paragraph{Analysis of Representation Space: PCA and Linear Evaluation} Analysis of the learned representations is conducted by PCA-based visual inspection and the linear evaluation. All the experimental cases are summarized in Table \ref{tab:cases_LE}. The visualization of the representations on the 2-dimensional PC space is shown in Fig. \ref{fig:pca_results} and the linear evaluation's test accuracy is shown in Table \ref{tab:cases_LE}. Note that the visual linear-separability by PCA would not always be matched with the linear evaluation results since PCA is an unsupervised method, meaning that the PCA's principal component axes are fitted without consideration of the labels. The PCA visualization shows that $Z_{\mathrm{TF-IDF}}$ seems to be linearly separable to some extent, however, its representations do not have good clustering but are rather scattered out. The linear evaluation results with respect to $Z_{\mathrm{TF-IDF}}$ are somewhat matched with the visual inspection by showing the test accuracy of around 89\%. $Z^\prime$ by Word2Vec, and GloVe, and BERT (word emb. only) show good clustering in the PCA visualization but the two clusters have some overlapping region which would discourage the linear separability. The linear evaluation shows the test accuracy of around 89\% for $Z^\prime$-s. Finally, $Z^\prime_{B}$ shows the good clustering and separability in the PCA visualization and its linear evaluation results in the highest test accuracy, reaching around 97\%. This result indicates that the contextualized word embedding where the contextualization is achieved by the BERT's transformer encoder significantly matters. Given that the model (h) could reach 97\% by training a single linear layer only, the performance is expected to be further improved by fine-tuning some part of the BERT's transformer encoder (\textit{e.g.,} a few top blocks of the transformer encoder).

\begin{table}[ht!]
\centering
\def\arraystretch{1.2}
\begin{tabular}{ ll c c c c c }
\hline
 & Vectorizer & \thead{Notation of \\ representation} & \thead{n-gram \\ type} & \thead{Stop word \\ removal} & \thead{Dimension \\ size} & \thead{Linear evaluation \\ (test acc.)} \\
\hline
(a) & TF-IDF                & $Z_{\mathrm{TF-IDF}}$ & unigram & yes & 300 & 0.892 \\
(b) & TF-IDF                & $Z_{\mathrm{TF-IDF}}$ & unigram & no  & 300 & 0.901 \\
(c) & TF-IDF                & $Z_{\mathrm{TF-IDF}}$ & bi-gram & yes & 300 & 0.891 \\
(d) & TF-IDF                & $Z_{\mathrm{TF-IDF}}$ & bi-gram & no  & 300 & 0.887 \\
(e) & Word2Vec              & $Z^\prime$    &         & no  & 300 & 0.888 \\
(f) & GloVe                 & $Z^\prime$    &         & no  & 300 & 0.89  \\
(g) & BERT \small{(word emb. only)}         & $Z^\prime$    &         & no  & 768 & 0.886 \\
(h) & BERT \small{(all embs. + encoder)} & $Z^\prime_B$ &         & no  & 768 & \textbf{0.969} \\ 
\hline
\end{tabular}
\caption{Summary of all the experimental cases. Note that $Z^\prime$ and $Z^\prime_B$ are static and contextualized word embeddings, respectively. "BERT (word emb. only)" denotes a vectorizer of the pre-trained word embedding matrix by BERT and the pooling layer while "BERT (all embs. + encoder)" denotes a vectorizer of all the pre-trained embeddings by BERT, the BERT encoder, and the pooling layer. For the test accuracy of the linear evaluation, the bold font denotes the best performance.}
\label{tab:cases_LE}
\end{table}

\begin{figure}[ht!]
\centering
\includegraphics[width=1.0\textwidth]{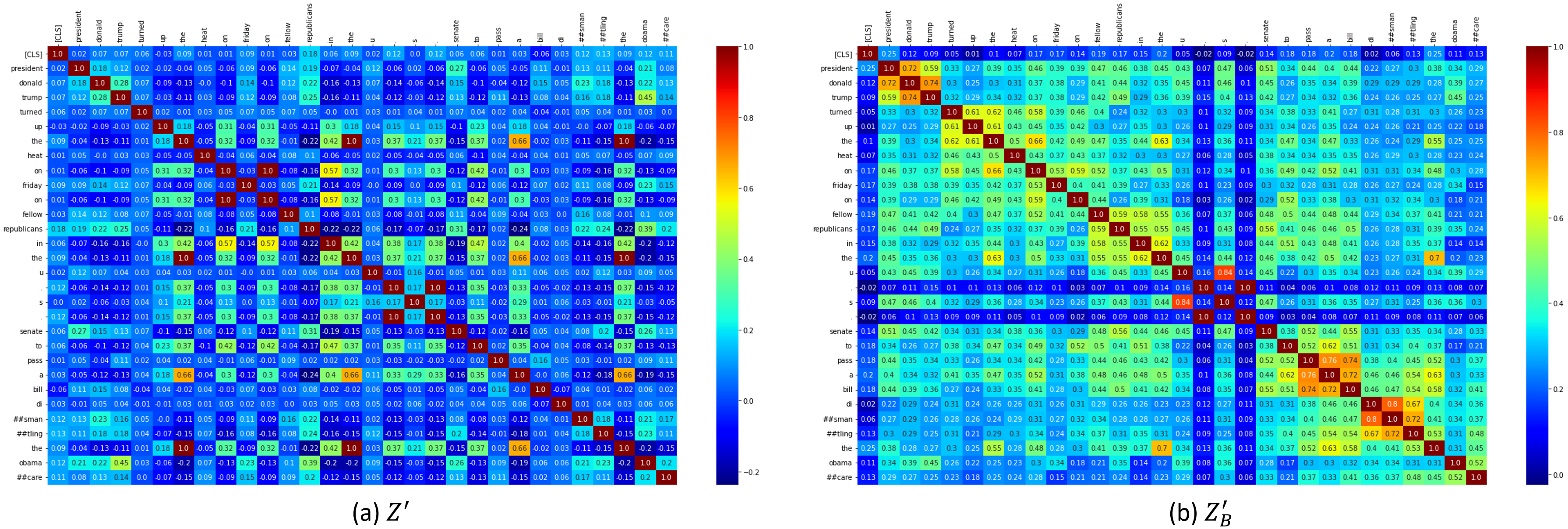}
\caption{Correlation matrices on $Z^\prime$ and $Z^\prime_B$ to analyze difference between the two representations along the temporal dimension and another difference imposed by the global temporal dependency encoded by the BERT's transformer encoder. The words can be clearly visible when zoomed in. The used sentence here is "president donald trump turned up the heat on friday on fellow republicans in the u.s. senate to pass a bill dismantling the obamacare". Note that the presented word tokens are words processed by the BERT's tokenizer.}
\label{fig:corr_Z}
\end{figure}

\paragraph{Analysis of Representation Space: $Z^\prime$ vs. $Z^\prime_B$}
In this paragraph, $Z^\prime$ by BERT and $Z^\prime_B$ right before the pooling layer are considered to analyze difference between the two representations along the temporal dimension and another difference imposed by the global temporal dependency encoded by the BERT's transformer encoder. Then, $Z^\prime$ and $Z^\prime_B$ have a dimension of $(B \times L \times D)$. The main difference between $Z^\prime$ by BERT and $Z^\prime_B$ is the contextualization process. Since the contextualization is achieved by the self-attention modules in the transformer encoder, it can be expected that tokens in $Z^\prime_B$ would have a higher correlation with other tokens with some semantic relations than $Z^\prime$. To verify the assumption, one sample is randomly selected from the test set, and correlation matrices on $Z^\prime$ and $Z^\prime_B$ are obtained and presented in Fig. \ref{fig:corr_Z}. The tokens of the entire sentence are truncated for the visualization. In Fig. \ref{fig:corr_Z}, several major patterns can be found: 1) overall correlations are higher in $Z^\prime_B$ due to the transformer's self-attention module, 2) neighboring tokens have high correlations due to the positional encoding, 3) In $Z^\prime$, the same word tokens have a correlation of 1.0 while the same word tokens do not have the perfect correlation in $Z^\prime_B$ this is because $Z^\prime_B$ involves both the positional encoding and the transformer encoder, which can encode the word relations in a context-aware manner. Some of the more detailed examples regarding the context-awareness are the correlation between u and s, correlation between u and s and donald and obama, and correlation between bill and \#\#care.

\begin{figure}[ht!]
\centering
\includegraphics[width=1.0\textwidth]{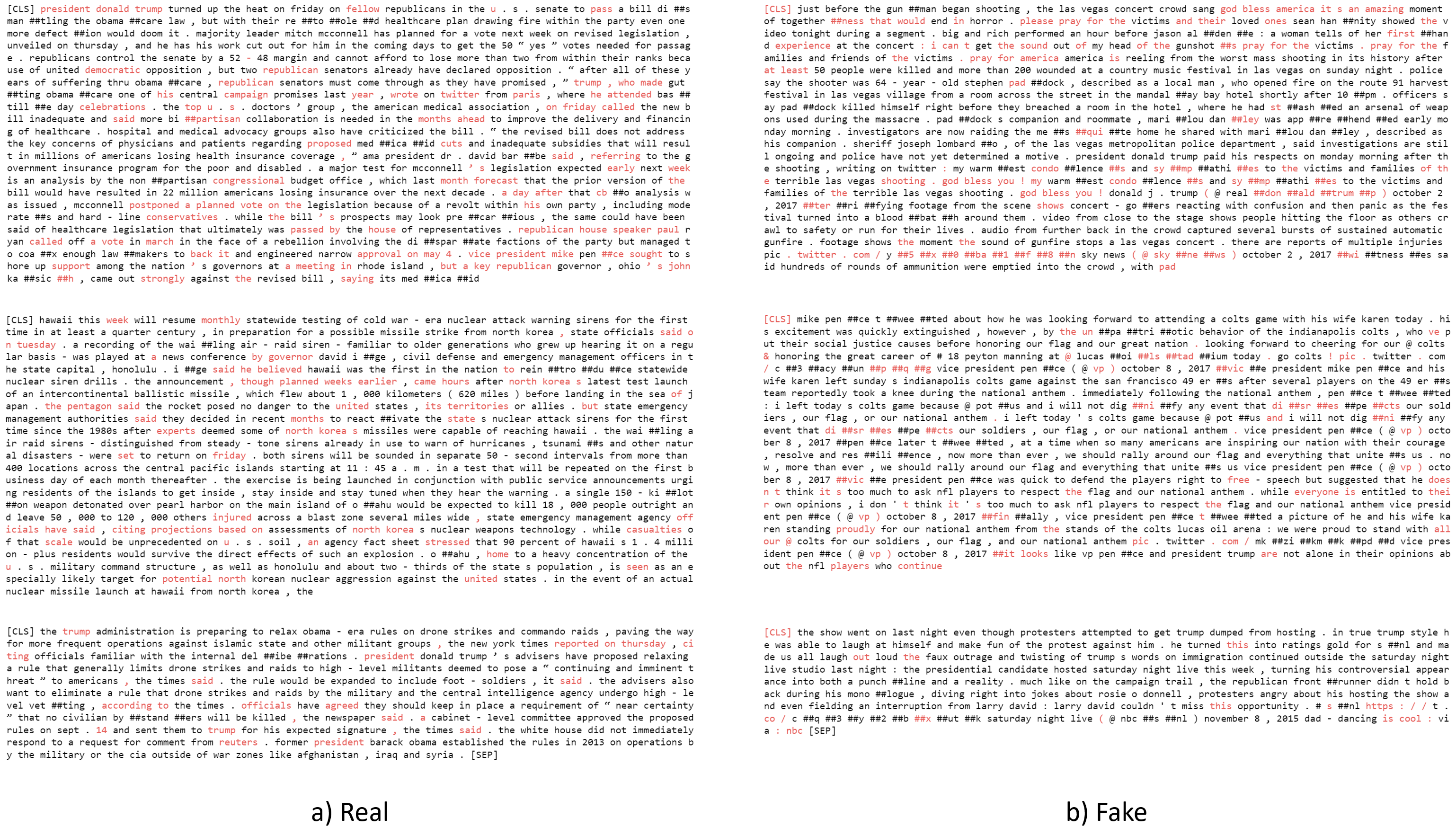}
\caption{Example sentences with the CAM. 10\% of words in a sentence with the highest $M_c(l)$ are colored in red. Note that the examples are randomly picked (\textit{i.e.,} no cherry picking). Long sentences are truncated to be fitted to the BERT encoder. Note that the presented word tokens are words processed by the BERT's tokenizer.}
\label{fig:cam_example_sents}
\end{figure}

\paragraph{CAM-based Interpretation Method for Pattern Discovery} In case of image data, $M_c$ can be visualized in continuous colors over different colors to represent continuous values of $M_c$. But, the visualization of $M_c$ on texts is quite tricky, therefore, 10\% of words in a sentence with the highest $M_c(l)$ are colored in red while the rest of the words are colored in black. Then, the red words are the words that are highly "focused" to classify a class $c$. Fig. \ref{fig:cam_example_sents} shows example sentences with the CAM along with their classes. Even by a quick look, the patterns of the real and fake articles seem obvious considering the focused words. For instance, the focus is made mostly on formal words and formal phrases for the real articles, and the focus is mostly on informal words, informal phrases, and punctuation marks such as :, (, ), /, @, \&, ! for the fake articles. Hence, the underlying pattern discovery for different classes through CAM is shown to be quite effective.

\section{Conclusion}
This study is about analysis of representation space, class-specific pattern discovery, and a simple classification model on the real and fake news dataset. The main contributions of this study are as follows: 1) analysis of the representation spaces by TF-IDF, Word2Vec, GloVe, and BERT through the PCA visualization and the linear evaluation, 2) application of CAM for class-specific pattern discovery, 3) achievement of both high classification accuracy and interpretability with a simple BERT-based architecture (\textit{i.e.,} naive BERT topped with a linear layer). CAM is shown to be used as a useful and simple tool to discover class-specific patterns so that one can have a better understanding of a predicted classification. The pre-trained BERT model with a single linear layer on top is shown to be robust and interpretable such that its classification test accuracy is high and the architecture is compatible with CAM. Lastly, a further study can be conducted by considering more datasets to investigate general capability of the proposals in this study.


\bibliographystyle{unsrt}  
\bibliography{references}

\end{document}